\documentclass[conference]{IEEEtran}
\IEEEoverridecommandlockouts
\usepackage{amsmath,amsfonts}
\usepackage{algorithmic}
\usepackage{array}
\usepackage{textcomp}
\usepackage{stfloats}
\usepackage{url}
\usepackage{verbatim}
\usepackage{graphicx}
\usepackage{orcidlink}
\usepackage{xcolor}
\usepackage{bookmark}
\usepackage{marvosym}
\usepackage{multirow}
\usepackage{booktabs}
\usepackage{makecell}
\usepackage{cite}
\usepackage{pifont}
\usepackage{hyperref}
\usepackage{authblk}
\usepackage{arydshln}

\def\BibTeX{{\rm B\kern-.05em{\sc i\kern-.025em b}\kern-.08em
    T\kern-.1667em\lower.7ex\hbox{E}\kern-.125emX}}
    
\hypersetup{
    colorlinks,
    linkcolor=blue,
    citecolor=blue,
    filecolor=black,
    urlcolor=blue
}
\usepackage{balance}
\title{HSRMamba: Efficient Wavelet Stripe State Space Model for Hyperspectral Image Super-Resolution}
\author[1, 2]{Baisong Li}
\author[1, 2*]{Xingwang Wang \thanks{*Corresponding Author}}
\author[1, 2]{Haixiao Xu}
\affil[1]{College of Computer Science and Technology, Jilin University}
\affil[2]{Key Laboratory of Symbolic Computation and Krowledge Engineering \cr
of Ministry of Education, Jilin University }
\affil[ ]{ \texttt{lbs23@mails.jlu.edu.cn, \{xww,haixiao\}@jlu.edu.cn}}

\begin{document}
\maketitle
\begin{abstract}
Single hyperspectral image super-resolution (SHSR) aims to restore high-resolution images from low-resolution hyperspectral images. Recently, the Visual Mamba model has achieved an impressive balance between performance and computational efficiency. However, due to its 1D scanning paradigm, the model may suffer from potential artifacts during image generation. To address this issue, we propose HSRMamba. While maintaining the computational efficiency of Visual Mamba, we introduce a strip-based scanning scheme to effectively reduce artifacts from global unidirectional scanning. Additionally, HSRMamba uses wavelet decomposition to alleviate modal conflicts between high-frequency spatial features and low-frequency spectral features, further improving super-resolution performance. Extensive experiments show that HSRMamba not only excels in reducing computational load and model size but also outperforms existing methods, achieving state-of-the-art results.
The open-source code is available at: \url{https://github.com/oldsweet/HSRMamba}.

\end{abstract}

\begin{IEEEkeywords}
Hyperspectral Image, Super-Resolution, State Space Model, Wavelet Transform, Vision Mamba
\end{IEEEkeywords}

\section{Introduction}
Hyperspectral images (HSI) have dozens or even hundreds of spectral bands, furnishing far richer spectral characteristics than standard RGB images and multispectral images (MSI). Consequently, hyperspectral images find extensive application in downstream tasks of computer vision, such as image classification, object detection. However, due to the limitations of imaging devices, a trade-off exists between the spatial and spectral resolutions of the captured hyperspectral images. Hyperspectral images typically possess relatively low spatial resolution, which severely hampers their further practical applications. The single hyperspectral image super-resolution technology (SHSR) aims to reconstruct a low-resolution hyperspectral image (LRHSI) into a high-resolution hyperspectral image (HRHSI), effectively addressing the issue of low resolution in hyperspectral images.
\begin{figure}[tp]
    \centering
    \includegraphics[width=1\linewidth]{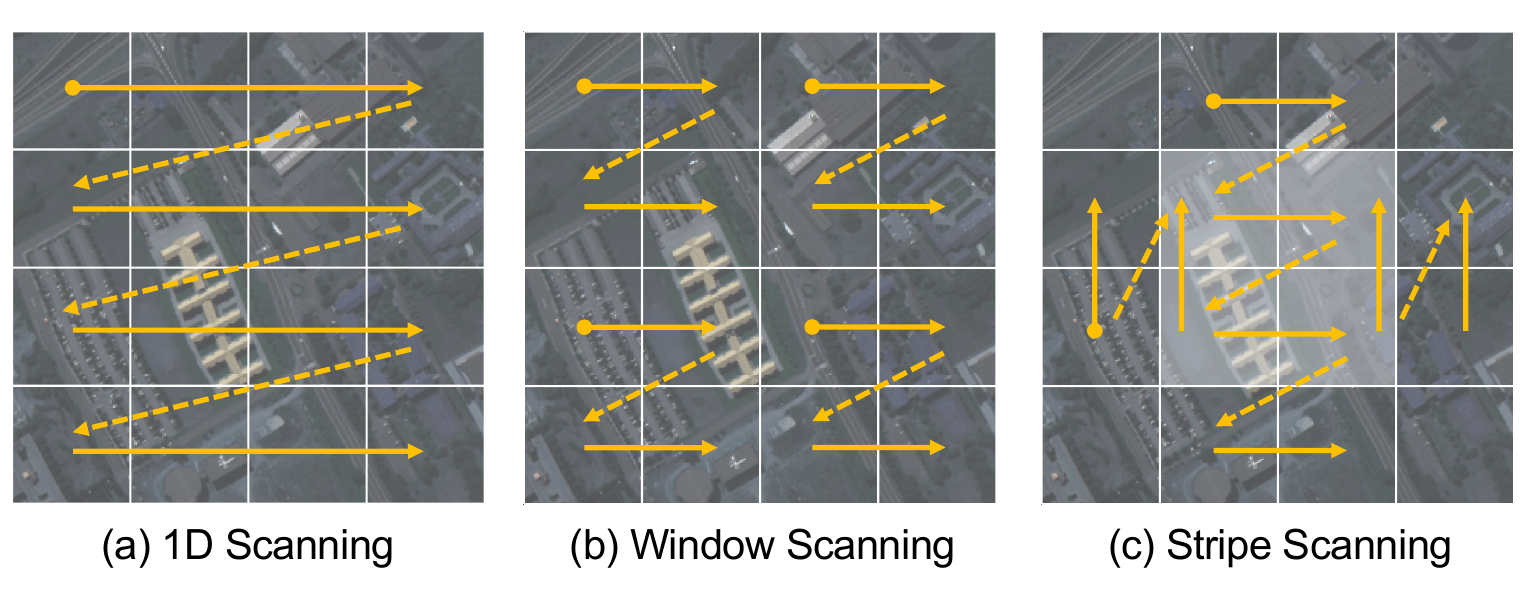}
    \caption{(a) 1D global selective scanning~\cite{MambaIR}. (b) window selective scanning ~\cite{windowMamba,localmamba}. (c) stripe 2D selective scanning in this work. Due to the presence of the cross-window, stripe scanning can more effectively capture both global and local features. }
    \label{fig:head}
\end{figure}

SHSR methods can be categorized into traditional methods and deep learning-based approaches. Traditional methods exploit the inherent properties of hyperspectral images, such as low rank and sparsity, but they exhibit weak generalization in complex scenarios. In recent years, deep learning-based SHSR methods have achieved remarkable progress. Convolutional Neural Networks (CNNs) \cite{SSPSR,DPIR,KSSANet} initially led breakthroughs in this field; however, their limited receptive field hinders the ability to capture long-range dependencies, thereby affecting reconstruction performance. To address the limitations of CNNs, researchers have introduced Transformer architectures~\cite{MSDFormer, ESSAformer}. Nevertheless, this introduces new challenges: first, Transformers require large-scale training data, which is difficult to obtain due to the scarcity of hyperspectral datasets; second, the computational complexity of the self-attention mechanism grows quadratically with the number of spectral tokens, resulting in a significant increase in computational cost.

State space models~\cite{ssm_01} have the advantage of performing global modeling of long sequences with linear complexity and have been widely applied in various tasks, particularly in scenarios that require efficient handling of long-range dependencies. Recently, the improved Mamba models have been successfully applied to several vision tasks and have demonstrated excellent performance~\cite{MambaIR, vmamba, vision_mamba, windowMamba,Zhu}. These models strike a satisfactory balance between performance and computational efficiency. However, due to the inherent 1D scanning paradigm of the visual Mamba models (as shown in Fig.~\ref{fig:head} (a) ), they are easily affected by image noise, leading to potential artifacts during the image generation process. Although some methods~\cite{windowMamba,localmamba} (such as the shown in Fig.~\ref{fig:head} (a)) attempt to improve this by incorporating window-based scanning patterns, these models still struggle to effectively capture global features. This is especially challenging in complex high-dimensional hyperspectral image scenarios, where the current visual Mamba models face limitations in capturing long-range dependencies and global information. Therefore, improving the scanning mechanism of visual Mamba models to effectively overcome noise interference, capture global features, and adapt to the characteristics of high-dimensional hyperspectral images remains a crucial problem that needs to be addressed.

To address these issues, as shown in Fig.~\ref{fig:head} (c), this paper proposes a new scanning method, which further improves the Visual Mamba model, named the Stripe Visual Mamba model, which introduces explicit cross-window~\cite{CSwin} interactions to balance global and local feature modeling. Furthermore, this paper proposes a novel hyperspectral image SR model named HSRMamba. To avoid conflicts between modeling low-frequency spectral features and high-frequency spatial textures during hyperspectral image reconstruction~\cite{wavemamba,vmambair}, HSRMamba employs a U-Net~\cite{Unet} architecture to effectively decompose and model high-frequency and low-frequency features in the wavelet frequency domain.

Specifically, based on the stripe visual Mamba model, we design three unique wavelet feature modeling modules.
1) To efficiently capture the low-frequency global spectral information components, we design the Low-Frequency Spectral Encoder (LFSE), which delves into the low-frequency features of spectral data in the wavelet domain.
2) To model the high-frequency spatial texture information, we design the High-Frequency Spatial Encoder (HFSE), which sensitively captures and processes the high-frequency spatial features of the image, accurately capturing the spatial texture details.
3) To fuse the features from two different modalities, we propose the High-Frequency and Low-Frequency Fusion Decoder (HLFD). This module organically combines the high-frequency and low-frequency features encoded by LFSE and HFSE, ensuring the comprehensive preservation of both spatial structure and spectral information in hyperspectral images.
The contributions of this paper are as follows:

\begin{itemize}
    \item This work presents HSRMamba, a novel SHSR model based on wavelet transform, which effectively integrates high-frequency spatial and low-frequency spectral features, achieving state-of-the-art performance.
    \item This study introduces an improved stripe-based Visual Mamba scanning method to mitigate artifact issues caused by image noise in traditional models.
    \item We develop three pluggable modules (LFSE, HFSE, and HLFD) for modeling and fusing low-frequency spectral and high-frequency spatial features.
\end{itemize}
\begin{figure}[!htb]
    \centering
    \includegraphics[width=1\linewidth]{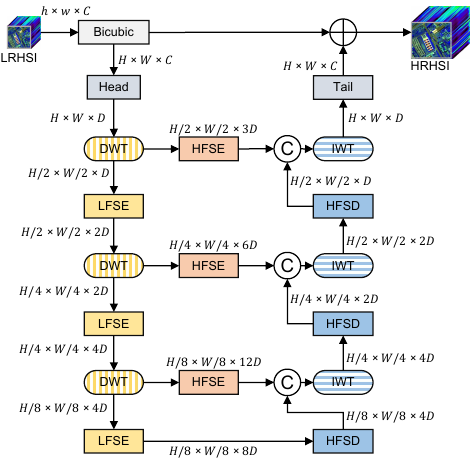}
    \caption{The overall structure of HSRMamba. DWT represents Discrete Wavelet Transform, and IWT represents Inverse Discrete Wavelet Transform.
    }
    \label{fig:overflow}
\end{figure}
\section{Proposed Method}
In this section, we introduce the overall structure of the proposed HSRMamba, and then present the implementation details of each sub-module.
\subsection{ Overall structure }
The overall structure is illustrated in Fig.~\ref{fig:overflow}. Given a LRHSI $ \mathbf{X} \in \mathbb{R}^{h \times w \times C} $, HSRMamba produces the corresponding high-resolution image $ \mathbf{Y} \in \mathbb{R}^{sh \times sw \times C} $, where $ h $, $w$, and $ C $ denote the height, width, and spectral channel count of the LRHSI, respectively, and $ s $ represents the spatial scaling factor.
The input LRHSI is first upsampled using Bicubic interpolation to generate $ \mathbf{X}^U \in \mathbb{R}^{sh \times sw \times C} $. Then, $ \mathbf{X}^U $ is passed through the Head module for shallow feature mapping, producing $\mathbf{X}^0 \in \mathbb{R}^{sh \times sw \times D} $.

In the downsampling stage, to eliminate the modeling modality conflict between low-frequency and high-frequency features, the feature $\mathbf{X}_i \in \mathbb{R}^{H \times W \times C} $ (where $i > 0$, $X_i$ represents the output of the $i$-th LFSE) is explicitly decomposed into high-frequency $\mathbf{X}_i^h \in \mathbb{R}^{H/2 \times W/2 \times 3C }$ and low-frequency components $ \mathbf{X}_i^l \in \mathbb{R}^{H/2 \times W/2 \times C} $  through \textit{discrete wavelet transform} (DWT).
The high-frequency feature $\mathbf{X}_i^h$ is fed into the HFSE module for high-frequency spatial feature modeling, while the low-frequency feature $\mathbf{X}_i^l$ is sent to the LFSE module for spectral feature modeling. In the upsampling stage, the high-frequency and low-frequency features are recombined through \textit{inverse discrete wavelet transform} (IWT) and subsequently fed into the HLFD module for the fusion of high-frequency and low-frequency features. Finally, the Tail module performs deep feature transformation to predict the residual values of the high-resolution image.
\begin{figure*}[!htb]
    \centering
    \includegraphics[width=0.9\linewidth]{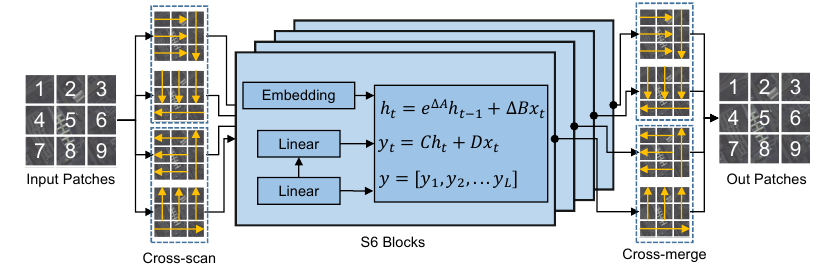}
    \caption{The architecture of the Strip 2D selective scanning module (stripe length is 2). S6 block is the state space model in~\cite{mamba}.}
    \label{fig:HSRMamba_strip_mamba}
\end{figure*}
\subsection{Stripe Visual Mamba}
Due to the inherent unidirectional scanning paradigm of Vision Mamba, vision models based on Mamba are susceptible to the influence of unidirectional image noise. Additionally, they struggle to learn local features effectively, which can introduce potential artifacts in the generated images. Although some efforts have been made to introduce window token scanning methods, these approaches significantly sacrifice global modeling capability of Mamba.

This paper proposes a new visual Mamba model, called Strip Visual Mamba, which fully integrates the advantages of both methods. Specifically, we use a stripe-based scanning approach to convert 2D images into token sequences as Fig.~\ref{fig:HSRMamba_strip_mamba}. This method inherently displays cross-window connections while partially capturing global features. For comparison purposes, we retain the structure of VSSM~\cite{MambaIR} but only replace the scanning method. The stripe length is fixed at 4. Notably, during the downsampling stage of HSRMamba, the fixed stripe length allows deeper features to obtain a larger local receptive field, which helps HSRMamba achieve more robust feature representation.

\subsection{Low-Frequency Spectral Encoder }
For the $i$-th LFSE, the low-frequency spectral feature $\mathbf{X}_{i-1}^l$ is fed into the LFSE module for spectral information modeling as Fig.~\ref{fig:HSRMamba_encoder}. Considering the sparsity of spectral features and to reduce computational complexity, the LFSE module first reduces the channel dimension of $\mathbf{X}_{i-1}^l$ to $\mathbf{X}_i^{l'}$ through a Head module.
To balance global and local spectral features, the LFSE adopts a dual-branch structure. One branch uses Channel Attention (CA)~\cite{SEnet} to capture local spectral features, while the other employs a VSSM module to extract global low-frequency spectral features. After passing through Sigmoid operation, the two features are multiplied using the Hadamard product. Finally, a residual operation is applied, and the channel dimension is restored through the Tail module and output $\mathbf{X}_{i}^l$ .

\subsection{High-Frequency Spatial Encoder}
Due to the limited receptive field of CNN models, capturing long-range high-frequency spatial correlations in images has been a significant challenge. HFSE combines multi-scale convolutions with global VSSM to effectively model local details while maintaining robust global feature representation.

\begin{figure}
    \centering
    \includegraphics[width=1\linewidth]{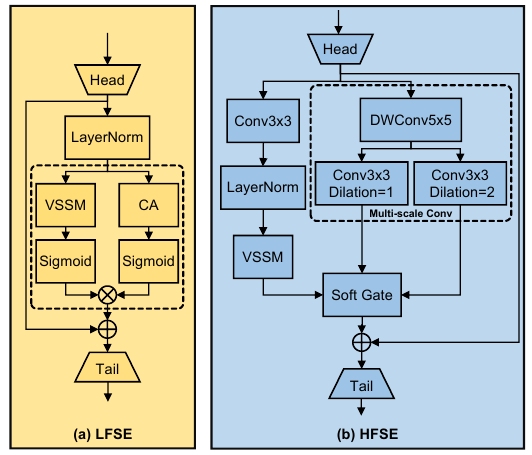}
    \caption{The structure of LFSE and HFSE. DSConv represents a $ 3 \times 3 $ depth-wise separable convolution. VSSM represents the Visual State Space Model in MambaIR~\cite{MambaIR} with stripe scanning. CA represents the Squeeze-and-Excitation operation~\cite{SEnet}, while Head and Tail refer to standard convolutions used for channel dimension mapping, with a channel scaling factor of 8.}
    \label{fig:HSRMamba_encoder}
\end{figure}

As shown in Fig.~\ref{fig:HSRMamba_encoder}, for the input $ \mathbf{X}_{i-1}^h $ of the $ i $-th HFSE module, it is first processed by the Head module to generate $\mathbf{X}_i^{h'} $. Next, the HFSE module processes global and local features through two separate branches. In the global feature processing branch, $ X_i^{h'} $ is first processed by a standard $ 3 \times 3 $ convolution, followed by LayerNorm normalization, and then processed by VSSM to extract global features $ \mathbf{X}_i^1 $. In the local feature processing branch, to address the limited receptive field of convolution, $ \mathbf{X}_i^{h'}$ is first processed by a $ 5 \times 5 $ depth-wise convolution, followed by two $ 3 \times 3 $ convolutions with different dilation rates, generating multi-scale local features $\mathbf{X}_i^{21}$ and $ \mathbf{X}_i^{22} $. Finally, the global and local features are fused using the \textit{soft gate}, and its mathematical expression is as follows:
\begin{equation}
    \mathbf{X}_i^{h'} = \frac{e^\alpha_i}{e^\alpha_i + e^{( 1 - \alpha_i )}} \cdot \mathbf{X}_i^{21} \cdot \mathbf{X}_i^1 +  \frac{e^{( 1 - \alpha_i )}}{e^\alpha_i + e^{( 1 - \alpha_i )}} \cdot \mathbf{X}_i^{22} \cdot \mathbf{X}_i^1
\end{equation}
$\alpha_i$ represents a trainable weight coefficient used to dynamically adjust the reference ratio of the global features. Finally, the updated $\mathbf{X}_i^{h'}$ is passed through a residual connection and then output $\mathbf{X}_i^{h}$  by the Tail module.

\subsection{High-Frequency and Low-Frequency Fusion Decoder}
Although inverse wavelet transform can explicitly fuse low-frequency and high-frequency features, direct fusion faces modal conflicts due to the inherent differences in neural network modeling. To efficiently enhance the fused features after inverse wavelet transform, this paper designs a lightweight and efficient fusion module called HLFD. As shown in Fig.~\ref{tab:main_res}, during the upsampling stage, for the $i$-th HFSE, the output $\mathbf{Y}_{i-1}$ after fusion of low-frequency and high-frequency features (where the first HFSE directly inputs the output of the last LFSE) is passed through the Head module to reduce the dimensionality, producing $\mathbf{Y}_{i}^{' }$. This output is then split along the channel dimension into two equal parts: $\mathbf{Y}_{i}^{1'}$ and $\mathbf{Y}_{i}^{2'}$, which are used for global and local feature modeling, respectively. The mathematical representation is as follows:
\begin{equation}
\left\{
    \begin{aligned}
    &\mathbf{Y}_{i}^{'} = Head(\mathbf{Y}_{i-1})     \\
    &\mathbf{Y}_{i}^{1'}, Y_{i}^{2'} = Split(\mathbf{Y}_{i}^{'})     \\
    &\mathbf{Y}_{i}^{1'} = VSSM(\mathbf{Y}_{i}^{1'}), \quad \mathbf{Y}_{i}^{2'} = DSConv_{3 \times 3}(\mathbf{Y}_{i}^{2'})                         \\
    &\mathbf{Y}_{i}^{'} = Conv_{3 \times 3}(Concat(\mathbf{Y}_{i}^{1'}, \mathbf{Y}_{i}^{2'})) + \mathbf{Y}_{i}^{'}, \\
    &\mathbf{Y}_{i} = Tail(\mathbf{Y}_{i}^{'})
\end{aligned}
\right.
\end{equation}
$Split$ and $Concat$ represent the splitting and concatenation along the channel dimension, respectively.

\section{Experiments}
To validate the effectiveness of HSRMamba, we conduct experiments on three publicly available hyerspectral image datasets.
The experimental details are as follows.

\subsection{Datasets and settings}
We perform comparative validation on three datasets: Chikusei, Pavia University (PaviaU), and Houston. Chikusei dataset consists of $2517 \times 2335$ pixels and contains 128 spectral bands. For training, we use the $1000 \times 2000$ region in the top-left corner of Chikusei, while the remaining part is divided into non-overlapping $680 \times 680$ regions for testing. PaviaU is an image with 103 spectral bands and a size of $610 \times 340$ pixels. The $340 \times 340$ region at the top is used for testing, with the remaining part used for training. The Houston image has a spatial size of $349 \times 1905$ pixels and contains 144 bands. We use the $349 \times 349$ region on the left side for testing, while the remaining region is used for training. In this study, we simulate LRHSI through Gaussian blurring (kernel size $3 \times 3$, standard deviation 0.5) and downsampling. When $ s = 2 $ or $ s = 4 $, the size of Ground Truth (GT) patch is 64, and when $ s = 8 $, the size of GT is 128. the hidden dimension $D$ fixed at 64.

We employ four widely used quality indicators (QIs) for quantitative testing: Peak Signal-to-Noise Ratio (PSNR), Spectral Angle Mapper (SAM), Relative Erreur Globale Adimensionnelle de Synthèse (ERGAS), and tructural Similarity Index (SSIM).

\subsection{Compared Methods}
To validate the effectiveness of HSRMamba, we compare it with six representative state-of-the-art models, including the Transformer-based ESSAFormer (ICCV'2023)~\cite{ESSAformer} and MSDFormer (TGRS'2023)~\cite{MSDFormer}, as well as the group convolution-based SSPSR (TCI'2020)~\cite{SSPSR}. For a broader comparison, we also evaluate models for RGB SR, including the lightweight U-Net convolutional model DPIR (TPAMI'2021)~\cite{DPIR}, the window attention-based SwinIR (ICCV'2021)~\cite{SwinIR}, and the standard image reconstruction model MambaIR (ECCV'2025)~\cite{MambaIR}, which is based on Visual Mamba. All methods are implemented using the PyTorch framework, with training conducted on an NVIDIA A800 GPU. The training task utilizes the $L1$ loss function, with an initial learning rate set to $ 1 \times 10^{-4} $, and AdamW as the optimizer. The batch size for training is 8, with a total of 50 epochs.

\begin{table*}[!htb]
\centering
\setlength\tabcolsep{4pt}
\caption{Quantitative comparison of different methods on the PaviaU, Chikusei and Houston datasets. The best values are highlighted in \textbf{bold}, and the second-best values are \underline{underlined}.}

\begin{tabular}{l|c|cccc|cccc|cccc}
\hline
\multirow{2}{*}{Methods} & \multirow{2}{*}{Scale} & \multicolumn{4}{c|}{PaviaU} & \multicolumn{4}{c|}{Chikusei} & \multicolumn{4}{c}{Houston}  \\
&& PSNR$\uparrow$  & SSIM$\uparrow$  & SAM$\downarrow$  & ERGAS$\downarrow$  & PSNR$\uparrow$  & SSIM$\uparrow$  & SAM$\downarrow$  & ERGAS$\downarrow$  & PSNR$\uparrow$  & SSIM$\uparrow$  & SAM$\downarrow$  & ERGAS$\downarrow$   \\
\hline
Bicubic            & $\times2$ & 30.1127 & 0.8624 & 4.1779 & 4.8890   & 36.0975 & 0.9389 & 3.1602 & 4.5915   & 32.9076 & 0.8777 & 6.3228 & 5.4270  \\
DPIR~\cite{DPIR}               & $\times2$ & \underline{32.6675} & 0.9087 & \underline{3.8956} & \underline{3.7149}   & \underline{40.7807} & \underline{0.9748} & \underline{2.4456} & \underline{2.9009}   & \underline{34.7586} & \underline{0.9166} & \underline{5.4862}& \underline{4.4041} \\
SSPSR~\cite{SSPSR}              & $\times2$ & 31.3265 & 0.8859 & 5.2062 & 4.3630   & 40.6419 & 0.9741 & 2.5210 & 2.9188   & 34.6908 & 0.9171 & 5.6028 & 4.5191 \\
MSDFormer~\cite{MSDFormer}  & $\times2$ & 32.6658 & \underline{0.9152} &4.2415&3.7546   & 39.8435 & 0.9681 & 2.9388 & 3.3252   & 34.5977 & 0.9139 & 5.6906 & 4.6105 \\
SwinIR~\cite{SwinIR}             & $\times2$ & 27.1588 & 0.7245 & 8.4923 & 7.1848   & 39.7435 & 0.9676 & 2.9165 & 3.4164         & 34.2063 & 0.9026 & 6.6013 & 4.9442 \\
ESSAFormer~\cite{ESSAformer}         & $\times2$ & 28.5603 & 0.8043 & 7.5993 & 6.3098   & 40.0782 & 0.9695 & 2.9545 & 3.4093   & 34.3545 & 0.9074 & 6.3988 & 4.8144 \\
MambaIR~\cite{MambaIR}            & $\times2$ & 28.4529 & 0.8124 & 7.1285 & 6.4460   & 40.2335 & 0.9720 & 2.9451 & 3.3363   & 34.3151 & 0.9085 & 6.2671 & 4.7683 \\
HSRMamba           & $\times2$ & \textbf{33.8863} & \textbf{0.9314} & \textbf{3.3939} & \textbf{3.2846}   & \textbf{40.8063} & \textbf{0.9749} & \textbf{2.4072} & \textbf{2.8322}   & \textbf{35.2265} & \textbf{0.9245} & \textbf{5.2524} & \textbf{4.1651} \\

\hline
Bicubic           & $\times4$ & 24.5756 & 0.6209 & 7.2560 & 9.2506   & 30.0468 & 0.8059 & 5.2342 & 9.0082   & 28.4500 & 0.7254 & 9.9584 & 9.1269  \\
DPIR~\cite{DPIR}               & $\times4$ & 26.4480 & 0.7253 & 6.6808 & 7.7807   & \underline{33.7416} & \underline{0.8955} & \underline{3.8457} & 6.2073   & \underline{30.8217} & 0.8033 & 8.3012 & \underline{6.9335} \\
SSPSR~\cite{SSPSR}              & $\times4$ & 27.1476 & 0.7323 & 6.7374 & 6.9973   & 33.7084 & 0.8917 & 3.8466 & \underline{6.1395}  & 30.8038 & \underline{0.8055}& \underline{8.0814} & 6.9443 \\
MSDFormer~\cite{MSDFormer}         & $\times4$ & \underline{27.4825}& \underline{0.7546} & \underline{6.1229} & \underline{6.7228} & 33.4451 & 0.8860 & 3.9393 & 6.3111   & 30.7886 & 0.7979 & 8.2971 & 6.9565 \\
SwinIR~\cite{SwinIR}              & $\times4$ & 24.6737 & 0.5538 & 10.4779 & 9.2714  & 33.1315 & 0.8749 & 4.3250 & 6.6253         & 30.6180 & 0.7907 & 8.8636 & 7.2065 \\
ESSAFormer~\cite{ESSAformer}        & $\times4$ & 25.9685 & 0.6732 & 8.6016 & 8.2817   & 33.6491 & 0.8887 & 4.1162 & 6.3759   & 30.6899 & 0.7945 & 8.7272 & 7.1551 \\
MambaIR~\cite{MambaIR}           & $\times4$ & 25.6164 & 0.6550 & 8.6770 & 8.6001   & 33.2209 & 0.8807 & 4.2432 & 6.6246   & 30.5883 & 0.7922 & 8.8003 & 7.1938 \\
HSRMamba          & $\times4$ & \textbf{27.6545} & \textbf{0.7555} & \textbf{5.9670} & \textbf{6.5293}  & \textbf{33.9321} & \textbf{0.8939} & \textbf{3.6627} & \textbf{6.1115}   & \textbf{30.9305} & \textbf{0.8076} & \textbf{8.0121} & \textbf{6.8639} \\

\hline
Bicubic           & $\times8$ & 21.2370 & 0.4752 & 12.2715 & 13.6723   & 26.9765 & 0.7185 & 7.2649 & 12.6484   & 25.4359 & 0.6272 & 14.9601 & 12.9911  \\
DPIR~\cite{DPIR}                 & $\times8$ & 22.0803 & 0.5253 & 12.2751 & 12.3510   & 28.6920 & 0.7551 & 6.1289 & 10.5824    & 27.2475 & 0.6841 & 13.2134 & 10.5566 \\
SSPSR~\cite{SSPSR}             & $\times8$ & \underline{22.8675} & \underline{0.5383} & 11.6572 & \underline{11.2788}   & 29.5184 & \underline{0.7892} &\underline{5.5789} & \underline{9.6007}   & \underline{27.3268} & \underline{0.6861} & \underline{12.8980}& \underline{10.4424}\\
MSDFormer~\cite{MSDFormer}          & $\times8$ & 22.7212 & 0.5172 & \underline{11.5969} & 11.4368   & 28.8613 & 0.7362  & 6.3057 & 10.3445   & 26.9969 & 0.6389 & 13.8842 & 10.8213 \\
SwinIR~\cite{SwinIR}           & $\times8$ & 21.1562 & 0.3254 & 17.2520 & 13.7513   & 29.4412 & 0.7778  & 5.9130 & 9.7471    & 26.9966 & 0.6684 & 13.8585 & 10.9708 \\
ESSAFormer~\cite{ESSAformer}         & $\times8$ & 22.4685 & 0.4857 & 13.1126 & 11.8328   & \underline{29.5837} & 0.7878  & 5.7324 & 9.6134    & 27.2077 & 0.6776 & 13.4150 & 10.6679 \\

MambaIR~\cite{MambaIR}           & $\times8$ & 21.9640 & 0.4679 & 13.5236 & 12.4221   & 28.9557 & 0.7563  & 6.3177 & 10.2179    &27.0085 & 0.6620 & 13.9308 & 10.8906 \\

HSRMamba         & $\times8$ & \textbf{23.1547} & \textbf{0.5892} & \textbf{11.1833} & \textbf{11.0358}   & \textbf{29.6141} & \textbf{0.7923}  & \textbf{5.4743} & \textbf{9.5294}    & \textbf{27.4841} & \textbf{0.6975} & \textbf{12.4907} & \textbf{10.3242} \\
\hline
         Best Value& -& $+\infty$&1&0&0 & $+\infty$&1&0&0 & $+\infty$&1&0&0 \\
\hline
\end{tabular}
\label{tab:main_res}
\end{table*}

\subsection{Qualitative Results}

Table~\ref{tab:main_res} presents the experimental results of three datasets at three different scale factors. The results show that HSRMamba outperforms existing methods across all three datasets and all scale factors, achieving the best performance in all QIs. Notably, although DPIR performs well in small scale SR, its performance deteriorates in large-scale SR tasks due to the limited receptive field caused by its fully convolutional architecture. In contrast, SSPSR and MSDFormer achieve excellent results in large scale SR, thanks to the effectiveness of global group convolution and self-attention mechanisms. Our proposed HSRMamba strikes an excellent balance between global and local feature modeling, achieving impressive results in both large-scale and small-scale SR tasks.

\begin{figure*}
    \centering
    \includegraphics[width=1\linewidth]{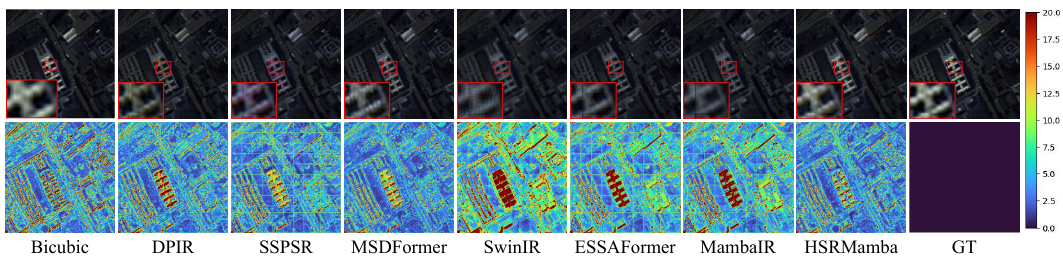}
    \caption{Pseudo-color images (R=20, G=30, B=40) generated by all comparative models for the testing area of the PaviaU dataset for $\times4$ SR and their corresponding SAM error maps.}
    \label{fig:HSRmamba_paviaU}
\end{figure*}

To showcase the subjective visual quality of the images generated by HSRMamba, we present the pseudo-color images and their corresponding SAM error maps generated on PaviaU datasets for $\times4$ SR in Fig.~\ref{fig:HSRmamba_paviaU}.
 It is evident that the image quality generated by HSRMamba surpasses that of other methods. The images produced by HSRMamba not only exhibit closer spatial details to the GT images, but the SAM error maps also demonstrate the stronger advantage of HSRMamba in spectral reconstruction performance. Furthermore, with the introduction of Stripe Mamba, HSRMamba significantly outperforms MambaIR, which relies on global 1D scanning, and effectively avoids potential artifacts.

\subsection{Efficiency Analysis}
To demonstrate our model's scalability and further proof of HSRMamba's efficiency and effectiveness, We report the running efficiency of each model on the PaviaU dataset for $\times4$ SR in Table~\ref{tab:HSRMamba_eff}. Compared to the second-best model, MSDFormer, HSRMamba significantly reduces the computational cost (\#FLOPs) by approximately 6.12 times, while decreasing the parameter (\#Param) count by about 6.09 times, and it outperforms MSDFormer in terms of image SR performance.
These results indicate that our wavelet-based stripe Mamba is a more efficient method.
\begin{table}[!htp]
    \centering
    \caption{Efficiency Comparison of Models on PaviaU Dataset for x4 SR.
    The best values are highlighted in \textbf{bold}, and the second-best values are \underline{underlined}.}
    \setlength\tabcolsep{4pt}
    \label{tab:HSRMamba_eff}
    \begin{tabular}{l|cc|cc}
\hline
    Model & \#FLOPs (G)& \#Param (M)  & PSNR$\uparrow$ & SSIM$\uparrow$\\
\hline
    DPIR~\cite{DPIR}             & 2.2649  & 8.0297 & 26.4480 & 0.7253  \\
    SSPSR~\cite{SSPSR}           & 2.7620 & \underline{0.9795}  & 27.1476 & 0.7323 \\
    MSDFormer~\cite{MSDFormer}   & 10.1776 & 13.8936 & \underline{27.4825} &\underline{ 0.7546 }\\
    SwinIR~\cite{SwinIR}         & \textbf{0.4706} & 1.8168  & 24.6737 &0.5538 \\
    ESSAFormer~\cite{ESSAformer} & 3.3338 & \textbf{0.8110}  & 25.9685& 0.6732 \\
    MambaIR~\cite{MambaIR}       & \underline{1.2752} & 3.6529  &25.6164&0.6550\\
    HSRMamba                     & 1.6223 & 2.2805  & \textbf{27.6545} &\textbf{0.7555}\\
\hline
    \end{tabular}
\end{table}

\subsection{Ablation Study}
To evaluate the impact of each component to the performance of HSRMamba, we conduct ablation experiments on the PaviaU $\times4$ SR dataset. First, we perform experiments based on the complete model architecture, and then gradually remove or replace key modules within the model to observe changes in performance.
First, we investigate the impact of the three core components: LFSE, HFSE, and HLFD.
Specifically, we replace the three core components with VSSM from MambaIR (keeping the Head and Tail to ensure a fair comparison) and retrain the model from scratch. The experimental results are shown in Table~\ref{tab:ablation_components}. After the replacement, the model's performance significantly decreases, while its computational cost and size increase. This indicates the clear advantages of the effectiveness and lightweight nature of the three core components we design.
\begin{table}[!htp]
    \caption{Ablation studies of different components on PaviaU Dataset for $\times4$  SR.}
    \setlength{\tabcolsep}{4pt}
    \label{tab:ablation_components}
    \centering
    \begin{tabular}{ccc|cc|cc}
    \hline
         LFSE & HFSE &  HLFD & \#Flops (G) & \#Param (M) & PSNR$\uparrow$ & SSIM$\uparrow$\\
    \hline
               \ding{55} & \ding{52}  & \ding{52} & 1.6414 & 2.4219 & 26.8771 & 0.7289 \\
               \ding{52} & \ding{55}  & \ding{52} & 1.6428 & 2.8639 & 27.0012 & 0.7304 \\
               \ding{52} & \ding{52}  & \ding{55} & 1.7066 & 2.8668 & 27.2287 & 0.7411 \\
               \ding{55} & \ding{55}  & \ding{55} & 1.7463 & 3.5916 & 25.2213 & 0.6350 \\
               \ding{52} & \ding{52}  & \ding{52} &  1.6223 & 2.2805 & 27.6545 & 0.7555 \\
    \hline
    \end{tabular}
\end{table}
\begin{figure}[!htp]
    \centering
    \includegraphics[width=1\linewidth]{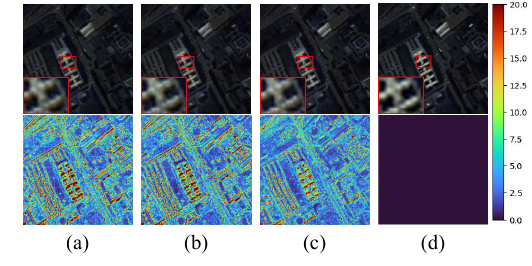}
    \caption{Pseudo-color images (R=20, G=30, B=40) generated by different Mamba scanning methods on the PaviaU dataset and their
corresponding SAM error map for $\times4$ SR. (a) Window Scanning. (b) Global Scanning.  (c) Stripe Scanning. (d) GT. }
    \label{fig:enter-label}
\end{figure}
To validate the effectiveness of the proposed stripe scanning method, we replace the scanning method in the three core components and retrain the model from scratch on the Pavia $\times4$ SR dataset. The experimental results are reported in Table~\ref{tab:ablation_scanning}, which show that the proposed stripe scanning method significantly outperforms the window-based and global scanning methods.
\begin{table}[!htp]
    \caption{Ablation studies on different selective scanning on PaviaU Dataset for $\times4$ SR. }
    \centering
    \begin{tabular}{l|cccc}
    \hline
    Scanning Method & PSNR$\uparrow$  & SSIM$\uparrow$  & SAM$\downarrow$  & ERGAS$\downarrow$  \\ \hline
    Window Scanning  & 26.5402  & 0.7303 & 6.6799 & 7.7801 \\
    Global Scanning  & 27.2911  & 0.7350 & 6.6013 & 6.8961 \\
    Stripe Scanning  & 27.6545  & 0.7555  & 5.9670 & 6.5293 \\
    \hline
    \end{tabular}
    \label{tab:ablation_scanning}
\end{table}
We further present the pseudo-color images generated by different scanning methods, as shown in Fig.~\ref{tab:ablation_scanning}. Although the window-based scanning method performs well in local detail texture and color restoration, it introduces window artifacts due to the lack of global information. The global scanning method avoids window artifacts but suffers from texture blurring and color distortion. Notably, the proposed Stripe scanning method achieves satisfactory results in both local detail and global feature restoration.
\section{Conclusion}
This paper presents a SHSR model based on the state space model. To avoid the modal conflicts between low-frequency spectral and high-frequency spatial features, HSRMamba introduces a network based on the U-Net architecture. Furthermore, three feature extraction and fusion modules are proposed to model and integrate high-frequency spatial texture features and low-frequency spectral features after wavelet decomposition. To address the potential artifacts caused by the 1D image scanning of the visual Mamba, this paper further improves the scanning method of the Visual Mamba.
\section{Acknowledgments}
This research was supported by the National Key Research and Development Program of China (No. 2023YFB4502304).
\bibliographystyle{IEEEtran}
\bibliography{main}

\begin{thebibliography}{10}
\providecommand{\url}[1]{#1}
\csname url@samestyle\endcsname
\providecommand{\newblock}{\relax}
\providecommand{\bibinfo}[2]{#2}
\providecommand{\BIBentrySTDinterwordspacing}{\spaceskip=0pt\relax}
\providecommand{\BIBentryALTinterwordstretchfactor}{4}
\providecommand{\BIBentryALTinterwordspacing}{\spaceskip=\fontdimen2\font plus
\BIBentryALTinterwordstretchfactor\fontdimen3\font minus \fontdimen4\font\relax}
\providecommand{\BIBforeignlanguage}[2]{{%
\expandafter\ifx\csname l@#1\endcsname\relax
\typeout{** WARNING: IEEEtran.bst: No hyphenation pattern has been}%
\typeout{** loaded for the language `#1'. Using the pattern for}%
\typeout{** the default language instead.}%
\else
\language=\csname l@#1\endcsname
\fi
#2}}
\providecommand{\BIBdecl}{\relax}
\BIBdecl

\bibitem{MambaIR}
H.~Guo, J.~Li, T.~Dai, Z.~Ouyang, X.~Ren, and S.-T. Xia, ``Mambair: A simple baseline for image restoration with state-space model,'' in \emph{European Conference on Computer Vision}.\hskip 1em plus 0.5em minus 0.4em\relax Springer, 2025, pp. 222--241.

\bibitem{windowMamba}
\BIBentryALTinterwordspacing
Z.~Wang, C.~Li, H.~Xu, X.~Zhu, and H.~Li, ``Mamba yolo: A simple baseline for object detection with state space model,'' 2024. [Online]. Available: \url{https://arxiv.org/abs/2406.05835}
\BIBentrySTDinterwordspacing

\bibitem{localmamba}
T.~Huang, X.~Pei, S.~You, F.~Wang, C.~Qian, and C.~Xu, ``Localmamba: Visual state space model with windowed selective scan,'' \emph{arXiv preprint arXiv:2403.09338}, 2024.

\bibitem{SSPSR}
J.~Jiang, H.~Sun, X.~Liu, and J.~Ma, ``Learning spatial-spectral prior for super-resolution of hyperspectral imagery,'' \emph{IEEE Transactions on Computational Imaging}, vol.~6, pp. 1082--1096, 2020.

\bibitem{DPIR}
K.~Zhang, Y.~Li, W.~Zuo, L.~Zhang, L.~Van~Gool, and R.~Timofte, ``Plug-and-play image restoration with deep denoiser prior,'' \emph{IEEE Transactions on Pattern Analysis and Machine Intelligence}, vol.~44, no.~10, pp. 6360--6376, 2021.

\bibitem{KSSANet}
B.~Li, X.~Wang, and H.~Xu, ``Kssanet: Kan-driven spatial-spectral attention networks for hyperspectral image super-resolution,'' in \emph{ICASSP 2025 - 2025 IEEE International Conference on Acoustics, Speech and Signal Processing (ICASSP)}, 2025, pp. 1--5.

\bibitem{MSDFormer}
S.~Chen, L.~Zhang, and L.~Zhang, ``Msdformer: Multi-scale deformable transformer for hyperspectral image super-resolution,'' \emph{IEEE Transactions on Geoscience and Remote Sensing}, 2023.

\bibitem{ESSAformer}
M.~Zhang, C.~Zhang, Q.~Zhang, J.~Guo, X.~Gao, and J.~Zhang, ``Essaformer: Efficient transformer for hyperspectral image super-resolution,'' in \emph{Proceedings of the IEEE/CVF International Conference on Computer Vision}, 2023, pp. 23\,073--23\,084.

\bibitem{ssm_01}
A.~Gu, I.~Johnson, K.~Goel, K.~Saab, T.~Dao, A.~Rudra, and C.~R{\'e}, ``Combining recurrent, convolutional, and continuous-time models with linear state space layers,'' \emph{Advances in neural information processing systems}, vol.~34, pp. 572--585, 2021.

\bibitem{vmamba}
Y.~Shi, M.~Dong, and C.~Xu, ``Multi-scale {VM}amba: Hierarchy in hierarchy visual state space model,'' in \emph{The Thirty-eighth Annual Conference on Neural Information Processing Systems}, 2024.

\bibitem{vision_mamba}
L.~Zhu, B.~Liao, Q.~Zhang, X.~Wang, W.~Liu, and X.~Wang, ``Vision mamba: Efficient visual representation learning with bidirectional state space model,'' \emph{arXiv preprint arXiv:2401.09417}, 2024.

\bibitem{Zhu}
C.~Zhu, S.~Deng, X.~Song, Y.~Li, and Q.~Wang, ``Mamba collaborative implicit neural representation for hyperspectral and multispectral remote sensing image fusion,'' \emph{IEEE Transactions on Geoscience and Remote Sensing}, vol.~63, pp. 1--15, 2025.

\bibitem{CSwin}
X.~Dong, J.~Bao, D.~Chen, W.~Zhang, N.~Yu, L.~Yuan, D.~Chen, and B.~Guo, ``Cswin transformer: A general vision transformer backbone with cross-shaped windows,'' in \emph{2022 IEEE/CVF Conference on Computer Vision and Pattern Recognition (CVPR)}, 2022, pp. 12\,114--12\,124.

\bibitem{wavemamba}
W.~Zou, H.~Gao, W.~Yang, and T.~Liu, ``Wave-mamba: Wavelet state space model for ultra-high-definition low-light image enhancement,'' in \emph{Proceedings of the 32nd ACM International Conference on Multimedia}, 2024, pp. 1534--1543.

\bibitem{vmambair}
Y.~Shi, B.~Xia, X.~Jin, X.~Wang, T.~Zhao, X.~Xia, X.~Xiao, and W.~Yang, ``Vmambair: Visual state space model for image restoration,'' \emph{arXiv preprint arXiv:2403.11423}, 2024.

\bibitem{Unet}
O.~Ronneberger, P.~Fischer, and T.~Brox, ``U-net: Convolutional networks for biomedical image segmentation,'' in \emph{Medical image computing and computer-assisted intervention--MICCAI 2015: 18th international conference, Munich, Germany, October 5-9, 2015, proceedings, part III 18}.\hskip 1em plus 0.5em minus 0.4em\relax Springer, 2015, pp. 234--241.

\bibitem{mamba}
A.~Gu and T.~Dao, ``Mamba: Linear-time sequence modeling with selective state spaces,'' \emph{arXiv preprint arXiv:2312.00752}, 2023.

\bibitem{SEnet}
J.~Hu, L.~Shen, and G.~Sun, ``Squeeze-and-excitation networks,'' in \emph{2018 IEEE/CVF Conference on Computer Vision and Pattern Recognition}, 2018, pp. 7132--7141.

\bibitem{SwinIR}
J.~Liang, J.~Cao, G.~Sun, K.~Zhang, L.~Van~Gool, and R.~Timofte, ``Swinir: Image restoration using swin transformer,'' in \emph{Proceedings of the IEEE/CVF international conference on computer vision}, 2021, pp. 1833--1844.

\end{thebibliography}
\end{document}